# Improving feature selection algorithms using normalised feature histograms

A. P. James and A. K. Maan

The proposed feature selection method builds a histogram of the most stable features from random subsets of training set and ranks the features based on a classifier based cross validation. This approach reduces the instability of features obtained by conventional feature selection methods that occur with variation in training data and selection criteria. Classification results on 4 microarray and 3 image datasets using 3 major feature selection criteria and a naive bayes classifier show considerable improvement over benchmark results.

*Introduction*: Relevance of features selected by conventional feature selection method is tightly coupled with the notion of inter-feature redundancy within the patterns [1-3]. This essentially results in design of data-driven methods that are designed for meeting criteria of redundancy using various measures of inter-feature correlations [4-5]. These measures behave differently to different types of data, and to different levels of natural variability. This introduces instability in selection of features with change in number of samples in the training set and quality of features from one database to another. The use of large number of statistically distinct training samples can increase the quality of the selected features, although it can be practically unrealistic due to increase in computational complexity and unavailability of sufficient training samples. As an example, in realistic high dimensional databases such as gene expression databases of a rare cancer, there is a practical limitation on availability of large number of training samples. In such cases, instability that occurs with selecting features from insufficient training data would lead to a wrong conclusion that genes responsible for a cancer be strongly subject dependent and not general for a cancer. This obviously means that the instabilities in feature selection, for example, when using gene expression databases are a problem of practical significance. In this Letter, we provide a novel method for reducing instability in the nature of features selected by using conventional feature selection methods through the new concept of creating normalised feature histograms from a given training set.

*Normalized feature histogram technique*: Figure 1 shows the various steps in the proposed feature histogram based feature selection algorithm. Statistically spread out training subsets are formed by several random selections of samples from a given training set. A conventional feature selection algorithm is applied to the individual training subsets to obtain a set of most relevant features and features are ranked by counting the number of times of their occurrence (a feature with higher count ranks higher than one with lower count). A histogram of ranked features arranged in the descending order of the ranks contains those features that are frequent and stable to variations in training set. The weights representing the individual ranked features are obtained by normalising the area of histogram to one. These weights indicate the probability of occurrence of individual ranked features when using a feature selection method on a given training set. The cumulative area obtained by adding the individual weights of features in descending order is used as a measure to globally select the stable set of features. The cumulative area at which the classification accuracy peaks is used as a cumulative area threshold. Figure 2 illustrates this approach to determine cumulative area threshold from maximum recognition accuracy and cumulative area distribution. Those features within cumulative area threshold have the maximum discriminative information, while the rest of the features are less useful for classification. Features are ranked in the order of their discriminatory ability by 10 fold cross-validation of individual features using a classifier to optimise ranked order of stable features with respect to classifier performance. The resulting ranked features are used for obtaining the classification accuracy on the testing set. It should be noted that the presented approach can be developed using any conventional feature selection criteria and provides a set of stable features that are less dependent to minor changes in training data. For example, the selection of a set of stable features in a microarray database irrespective of minor changes in training data is consistent with a boarder understanding that there are certain genes that are reflective of discriminating a disease/condition, as opposed to conventional feature selection methods that are less reflective of this issue.

*Experimental details*: We have used 4 benchmark microarray gene expression datasets GLI-85, CLL-SUB-111, TOX-171, SMK-CAN-187 [6] and 3 image datasets AR10P, PIX10P and ORL10P [6] to demonstrate our approach to feature selection. Individual datasets are divided randomly into training and testing sets of equal sizes. A random 20% of samples are chosen from the training set and conventional feature selection criteria [6] are applied to this training subset to obtain a set of top 100 relevant features. To demonstrate our idea we use 3 benchmark feature selection criteria listed in Table 1 (i.e. Chisquare, Fisher and Infogain [6]). Repeating random selection of training subset and feature selection for a statistically significant 100 times, a set of most relevant feature are obtained. Naive bayes classifier is used for cross validation tests on training test, and for studying classifier performance.

*Results and Discussion*: Table 1 demonstrates the significant improvement in classification accuracy by using the proposed method on 3 benchmark rank based feature selection criteria [6]. The accuracy comparison is reported for lowest number of features within the top 200 ranked selected features for which high accuracies are obtained. The maximum classification accuracy of conventional method with less than or same number of features to that of presented method are shown in Table 1. The average classification accuracies on individual datasets show an improvement of a maximum 6%. These results show that by using the proposed approach high recognition accuracies can be achieved with very small number of features. Further, we find from Table 1 that less than 200 features are sufficient to obtain high classification performance. Hence, this method can reduce the complexity of high dimensional datasets down to small number of important features while still improving classification accuracies. The robustness of the method is demonstrated by the fact that the normalised feature histogram method can be implemented on any available feature selection algorithm. Although only gene expression classification and image recognition are shown as examples, the method is general and practical to be applied for text mining, image retrieval, pattern recognition and other similar applications that require feature selection.

*Conclusion*: This Letter presented a method to improve the accuracy of conventional feature selection methods by determining a stable set of features from normalised feature histograms. Robustness and high classification accuracy is demonstrated for benchmark high dimensional gene expression and image datasets with substantial performance improvement over benchmark results.

**Acknowledgements**

The authors would like to acknowledge the support of Queensland Micro- and Nanotechnology Centre.

**Authors' affiliations:**

A. P. James and A. K. Maan (Queensland Micro- and Nanotechnology Centre, Griffith University, Qld. 4111)

E-mail: a.james@griffith.edu.au

**Figure captions:**

Figure 1 Normalised feature histogram based global feature selection and ranking method.

Figure 2 Illustration of cumulative area threshold determination and feature selection by maximum accuracy obtained in cross validation of training set.

Figure 1

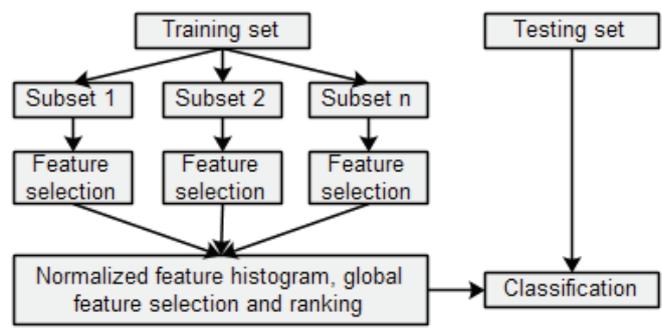

Figure 2

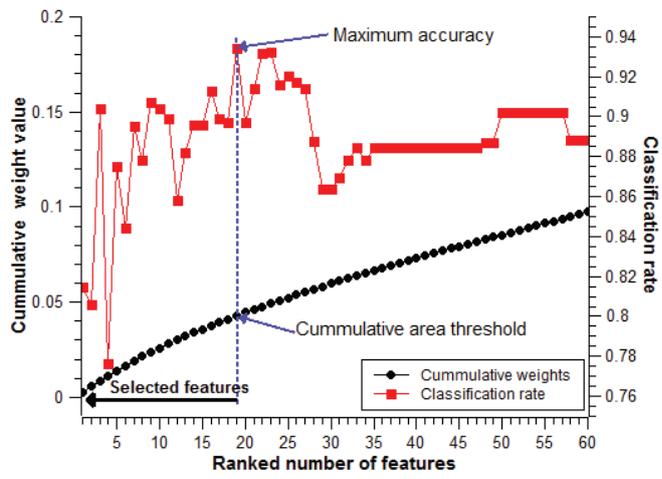

**Table captions:**

Classification performance comparison of proposed method in contrast to conventional method

Table 1

| Dataset | Method | Classification accuracy (%) | | | |
|---|---|---|---|---|---|
| | | Chisquare | Fisher | Infogain | Average |
| GLI-85 | **Presented** | **81.3±3.2** | **88.3±3.2** | **81.3±3.2** | **86.3±3.2** |
| | Conventional | 77.7±4.9 | 86.0±5.0 | 77.7±4.9 | 80.4±4.9 |
| CLL-SUB-111 | **Presented** | **70.7±4.3** | **64.2±4.5** | **71.4±2.5** | **68.7±3.7** |
| | Conventional | 64.6±4.9 | 61.3±5.9 | 67.8±5.2 | 64.5±5.3 |
| TOX-171 | **Presented** | **70.6±4.4** | **68.3±3.3** | **70.4±3.5** | **69.7±4.1** |
| | Conventional | 69.6±4.9 | 65.9±6.2 | 67.0±4.7 | 67.5±5.2 |
| SMK-CAN-187 | **Presented** | **70.0±3.9** | **66.8±4.9** | **70.3±3.6** | **69.0±4.1** |
| | Conventional | 65.6±3.9 | 65.9±4.8 | 65.9±2.9 | 65.8±3.9 |
| AR10P | **Presented** | **71.1±5.2** | **73.3±2.3** | **71.1±5.2** | **71.8±4.2** |
| | Conventional | 70.2±7.1 | 72.2±2.4 | 71.3±6.2 | 71.2±5.2 |
| PIX10P | **Presented** | **94.0±1.4** | **97.6±2.5** | **92.6±1.1** | **94.7±1.7** |
| | Conventional | 88.6±2.3 | 97.0±3.4 | 89.3±1.1 | 91.6±2.3 |
| ORL10P | **Presented** | **90.2±3.3** | **94.5±4.7** | **89.6±3.3** | **91.4±3.7** |
| | Conventional | 87.6±3.1 | 93.1±4.5 | 85.2±1.9 | 88.6±2.9 |